\pdfoutput=1
\documentclass[11pt]{article}
\usepackage{acl}
\usepackage{times}
\usepackage{latexsym}
\usepackage[T1]{fontenc}
\usepackage[utf8]{inputenc}
\usepackage{microtype}
\usepackage{graphicx} 
\graphicspath{ {./images/} }
\usepackage[utf8]{inputenc}
\usepackage{natbib}
\usepackage{latexsym}
\usepackage{graphicx}
\usepackage{soul}
\usepackage{hyperref}
\usepackage{array}
\usepackage{amssymb}
\usepackage{algorithm} 
\usepackage{algpseudocode}
\usepackage{tabulary}
\usepackage{dirtytalk}
\usepackage{pgfgantt}
\usepackage{xcolor}
\usepackage[utf8]{inputenc}
\usepackage{amssymb}
\usepackage{pifont}
\usepackage{titlesec}
\usepackage[utf8]{inputenc}
\usepackage[english]{babel}
\usepackage{array}
\usepackage{textcomp}

\title{An Empirical Study of Topic Transition in Dialogue}

\author{
  \begin{tabular}{c}
    Mayank Soni* 
  \end{tabular}%
  
  \begin{tabular}{c}
    Brendan Spillane* \\
  \end{tabular}
  \begin{tabular}{c}
    Emer Gilmartin\textsuperscript{†}
  \end{tabular}\\ \\
  \begin{tabular}{c}
    \textbf{Christian Saam\textsuperscript{†}}\\
  \end{tabular}
  \begin{tabular}{c}
  \textbf{Benjamin R. Cowan \textsuperscript{‡}}\\
  \end{tabular}
  \begin{tabular}{c}
    \textbf{Vincent Wade*}
  \end{tabular}\\[10pt]
  
  \begin{tabular}{c}
        *ADAPT Centre, Trinity College Dublin, \textsuperscript{†}Trinity College Dublin, \textsuperscript{‡}ADAPT Centre, University College Dublin \\
       \{sonim, spillab, saamc, gilmare, vincent.wade\}*@tcd.ie, 
       \{benjamin.cowan\}\textsuperscript{‡}@ucd.ie
       
  \end{tabular}

  }

\begin{document}
\maketitle
\begin{abstract}

Transitioning between topics is a natural component of human dialog. Although topic transition has been studied in dialogue for decades, only a handful of corpora based quantitative studies have been conducted to investigate the nature of topic transitions. Towards this end, this study annotates $215$ conversations from the switchboard, perform quantitative analysis and finds that longer conversations have more topic transitions, topic transition is generally carried out more by one participant and there is no pattern observed in time series of topic transition. This work presents an empirical study on topic transition in switchboard corpus followed by modelling topic transition with a precision of $91\%$ for in-domain($id$) test set and $78\%$ on $10$ \textit{out-of-domain}($ood$) test set. It is envisioned that this work will help in emulating human-human like topic transition in open-domain dialog systems. 
\end{abstract}


\section{Introduction}

Human conversation consists of multiple natural topic transitions, from introductions, to topics of interest, and on to leave talking, and thus relies on topic change and shading mechanisms to allow participants to maintain and change topics\footnote{Our annotated dataset and models do not differentiate between the types of topic transition (change, shift, shading, fading etc.) depicted in Gardner's model \cite{Gardner_1984}. For simplicity, this paper uses `topic transition' to describe all forms. Where necessary, it uses specific terms to differentiate.}. An example of topic transition can be seen in Figure 1, participants first begin by talking about each others age, then move on to the places participants want to visit and finally moving on to talking about the state of Arizona in the USA.
Although topic transition has been studied in linguistics for decades \citep{Gardner_1984, Lambrecht_1996, riou2015methodology, Van_Dijk_1977}, there are only a few corpora based studies investigating the nature of topic change. This is because of the labour intensive task of manually annotating datasets. Even though the task of annotation is labour intensive and manual, it is necessary to empirically understand how human participants engage in topic transition in a conversation. Towards this end, this work annotates $215$ conversations from the Switchboard \cite{Godfrey_Holliman_1993} corpus and studies different aspects of topic transition. To the authors best knowledge, this is the the largest quantitative study conducted on the nature of topic transition in social conversations till date.

\begin{figure} [htbp!]
    \begin{flushleft}
    \includegraphics[width=0.51\textwidth]{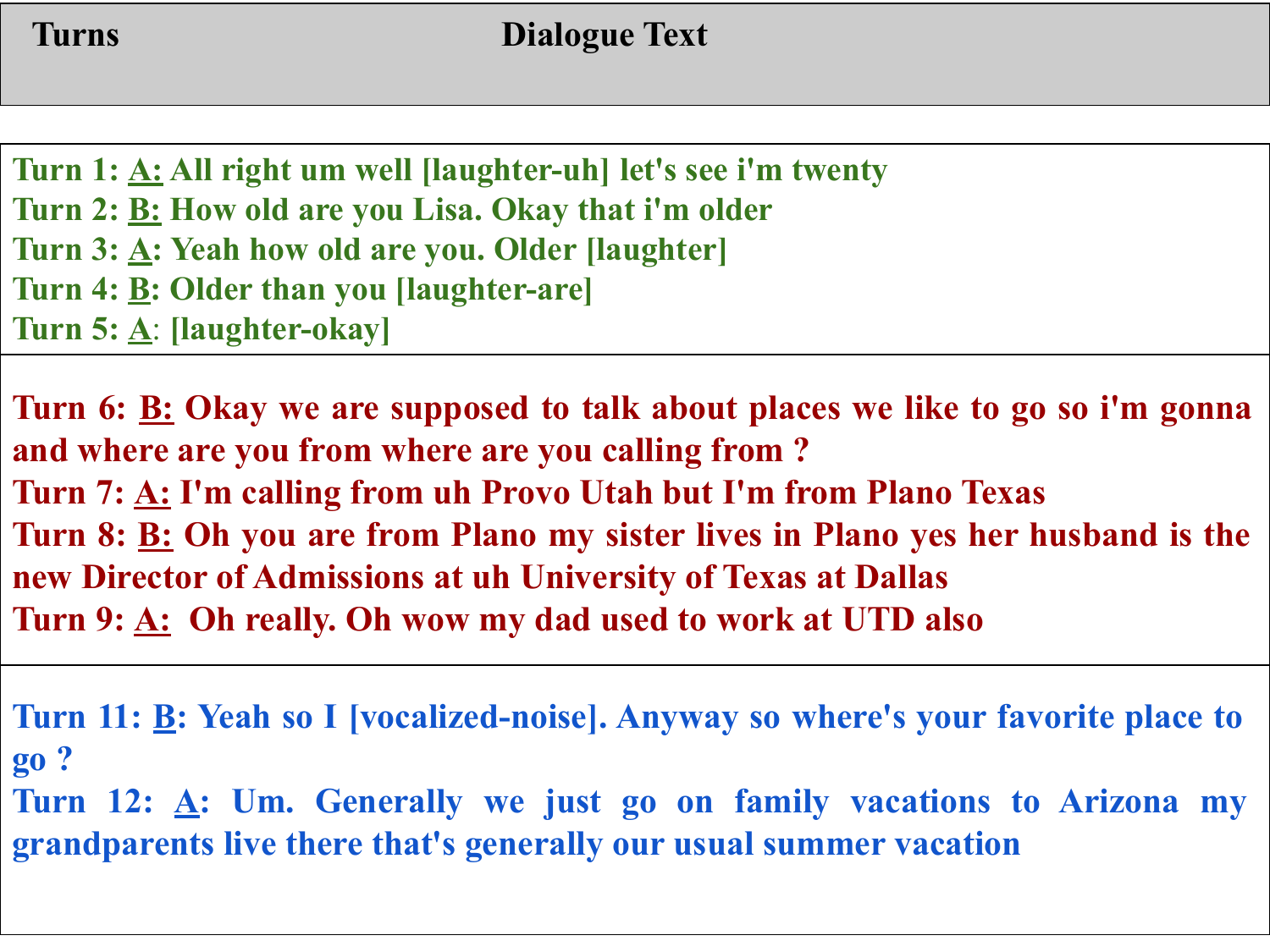}
    \caption{Hand-picked example of topic transition in the Switchboard corpus}
    \end{flushleft}
\end{figure}


Below, we explore the motivation, background and related work. We then describe empirical findings in the corpus, followed by modelling topic transition and evaluating it on in-domain ($id$) and out-of-domain ($ood$) test sets.


\section{Background Theory}
Definitions of topic in the literature fall into two categories; sentence level \citep{Lambrecht_1996} and discourse level \citep{Van_Dijk_1977}. As this study is concerned with discourse level topic annotation, we adopt the definition of \citet{Bonin_Campbell_Vogel_2012} which maintains that topic at a discourse level is the \textit{"segments of the discourse sharing coherent information (about the same thing)"}.

Topic transition has been categorized by \citet{Gardner_1984}, whose model of topic development in spoken interaction details the multiple means by which humans introduce, maintain, and change topics. Two areas which have received particular attention in the literature are topic change and topic shift. They have been defined as the point between two pieces of discourse which are considered to have different topics. \citet{Bublitz_1988} differentiates between topic change and topic shift as having low and high degrees of connectivity respectively to the previous topic. Topic shift includes both topic shading and topic fading \citep{Maynard_2009,Brown_Yule_1983,Garcia_Joanette_1997}. Topic change includes reintroduction and full blown change. We annotate all such topic transitions under one common label.


\section{Related Work}
Related work in the literature is primarily found in the domains of \textit{manual topic annotation} and \textit{automatic topic segmentation}. The basic approaches in both of these areas can be divided into manual and automatic. 


\subsection{Manual Annotation or Segmentation}
Early work to manually annotate topic transition was predominantly done for the purpose of conversation analysis. \citet{Planalp_Tracy_1980} were among the first to annotate topic transition. They showed that information integration by the interlocutors impacts their topic transition strategies. \citet{Crow_1983}\textquotesingle s analysis of topic shift in couples' conversations showed that it occurred fairly frequently; every 48 seconds on average. Later work by \citet{Ries_2001} showed that speaker initiative and style can also be indicative of topic transition. Recently, \citet{konigari2021topic} annotated a subset of the switchboard corpus \cite{Godfrey_Holliman_1993} into \emph{major}, \emph{minor} and \emph{other} topics. \citet{sevegnani2021otters} introduced a one-turn topic transition corpus by asking annotators to produce bridging sentence connecting two sentences of different topics. More recently, automatic annotation or segmentation has increasingly become the focus of research due to the significant costs and effort of manual annotation. 


\subsection{Automatic Segmentation}
There have been many attempts to automatically annotate or segment text based on topic. A detailed overview of early work is provided by \citet{Purver_Tur_De_Mori_2011}. There have been two approaches : applying algorithms that work with text segmentation to dialog segmentation and algorithms specifically invented to model dialog topical transitions. These approaches can also be divided into unsupervised and supervised approaches. Among the earliest relevant works is that of \citet{reynar1994automatic} who proposed a method of identifying topic boundaries based on lexical cohesion and dot plots. \citet{hearst1997texttiling} developed an unsupervised method to separate texts into multiple paragraphs representing subtopics. \citet{Passonneau_Litman_1997} developed two algorithms that use utterance features to segment dialogue by topic. \citet{Boufaden_Lapalme_Bengio_2001} used Hidden Markov Models to segment transcriptions of telephone conversations into topics. \citet{Galley_McKeown_Fosler_Lussier_Jing_2003} tackled the difficult problem of topic segmentation in multiparty speech by focusing on the content of the transcripts and their form, \emph{i.e.} the linguistic cues in the speech. \citet{Hsueh_Moore_Renals_2006} built on the work of \citet{Galley_McKeown_Fosler_Lussier_Jing_2003} by combining Automatic Speech Recognition (ASR) with existing text based methods of topic segmentation. \citet{Arguello_Rose_2006} also adopted a hybrid approach by combining linguistic features with local context indicators in the text. \citet{Sapru_Bourlard_2014} demonstrated that latent topic features are effective predictors of topic transition in transcripts of multiparty speech from office meetings. \citet{joty2011supervised} developed a supervised method of segmenting topic in email conversations. More recently, \citet{zhang2019topic} introduced a method based on BERT and TCN (Temporal Convolution Network). \citet{xing2021improving} introduced an unsupervised method for topical segmentation of dialog by utterance-pair scoring. There are other relevant techniques and we skip them in the interest of brevity.


\section{The Annotation Framework}
We annotate $215$ conversations from the Switchboard-1 Release $2$ corpus \citep{Godfrey_Holliman_1993}. Before describing the annotation framework, we briefly describe the Switchboard Corpus \cite{Godfrey_Holliman_1993} below in section 4.1. 

\subsection{The Switchboard Corpus }
The Switchboard-1 Release $2$ Corpus \citep{Godfrey_Holliman_1993} consists of recordings of about $2400$ telephone conversations between $543$ distinct speakers who did not know each other \citep{Calhoun_Carletta_Brenier_Mayo_Jurafsky_Steedman_Beaver_2010}. All interlocutors spoke American English. They choose a topic from a list of about $70$ topics and were connected to another interlocutor by a switchboard robot. About $50$ of the $70$ topics were chosen regularly. The conversation is not limited to the initial topic and participants could transition topics at any time. The individual conversation transcripts have been transcribed and annotated to the utterance level and include conversation $ID$s, time stamps, and label for speakers identity. 

\subsection{Annotation Framework}
For empirically studying topic transition and modelling topic transition $215$ conversations, drawn at random from the switchboard corpus are annotated. The annotation were performed for start \emph{(S)} and end \emph{(E)} of the conversation, greeting and leave taking \emph{(GIL)}, topic, topic transition \emph{(C)}, and failed topic transition \emph{(X)}. This manually annotated corpus consists of $20,566$ turns from $215$ conversation. Table 1 displays the statistics of annotated dataset. The average number of turns per conversation is $96$ with the shortest conversation lasting $33$ turns and the longest conversation lasting $242$ turns. Annotations are based on previous studies demonstrating that naive annotators are capable of annotating topic transition with success \citep{Mann_Carlisle_Moore_Levin_1977,Passonneau_Litman_1997,Planalp_Tracy_1980}. The conversations were annotated by two annotators. The inter-annotator agreement (Cohen \textquotesingle s Kappa) obtained on a sample of five conversations is $0.64$, signifying substantial agreement.

\newcolumntype{M}[1]{>{\raggedright\arraybackslash}m{#1}}
\begin{table}[htbp!]
\scalebox{01}{
    \footnotesize
    \raggedleft
    \begin{tabular}{
    M{0.7\linewidth} |                
    M{0.1\linewidth}}

    \\
{No. conversations} & 215
 \\ \hline \\ 
{No. total turns} & 20,566
 \\ \hline \\ 
{Avg. turns per conversation} & 96
 \\ \hline \\
{Avg. topics transitions per conversation} & 8
 \\ \hline \\
{Avg. turns per topic transitions} & 12
 \\
 
\end {tabular}}
\caption{Annotated switchboard dataset statistics}
\end{table}

\section{Empirical Studies of Topic Transitions}

Having obtained an annotated corpus of $215$ corpus, we conducted quantitative analysis on some aspects nature of topic change. The empirical findings are discussed in the subsections below.


\subsection{Relationship Between Length of A Conversation and Number of Topic Transitions}

\begin{figure} [htbp!]
     \begin{flushleft}
     \includegraphics[width=0.48\textwidth]{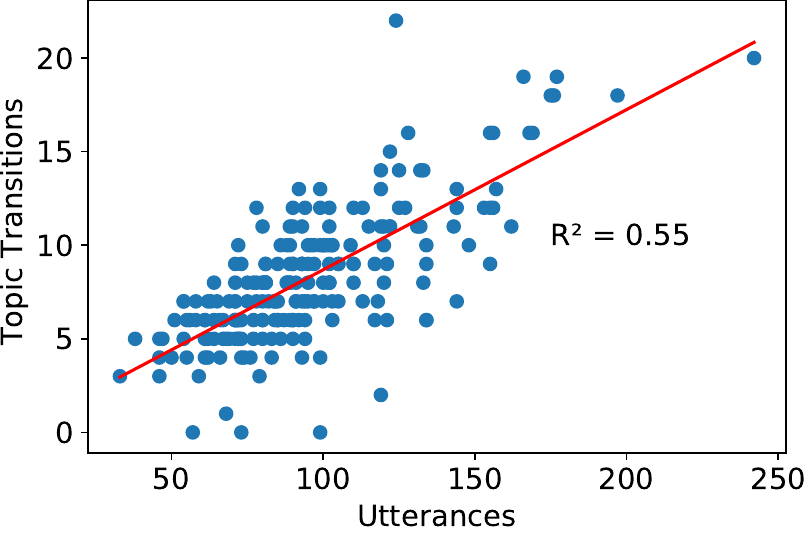}
     \caption{Scatter plot of number of topic transitions and length of conversations}
     \end{flushleft}
\end{figure}

Longer conversation are a sign of successful and engaging conversation. We wanted to examine if longer conversation consist of more topic transitions than shorter conversation or the number of topic transitions remains similar and some topics are conversed for more turns than others. Towards investigating this relationship, we calculate number of topic transitions per conversation and plot it in Figure 2. 

The Pearson correlation coefficient observed is observed to be $0.74$, indicating a positive correlation between length of a conversation and number of topic transitions. We also plot a linear regression line and observe a $R^2$ value of $0.55$ ($p << 0.001$). Figure 2 further highlights that number of topic transitions increase as length of a conversation increases. Most conversations consist of five to thirteen topic transitions. Thus, it is observed that longer conversations have more topic transitions. 

\subsection{Share of Topic Transition by Participants}
We wanted to explore further if the topic transitions are carried out evenly by both participant or if, one participant carries out more topic transitions. To investigate this, we first calculate the difference in number of topic transitions carried out by each participant for each conversation. We observe that only about 38\% of conversations had an equal or only one more topic transition than the other per participant. In about 62\% of conversations, one participant initiated at least two more topic transition than the other. Figure 3 shows a bar plot of topic transition difference and percentage of conversations with such difference. It is thus observed that topic transitions are unequally carried out between participants $(\tilde{\chi}^2 = 403.41, p << 0.005)$. 

\begin{figure} [htbp!]
     \begin{flushleft}
     \includegraphics[width=0.49\textwidth]{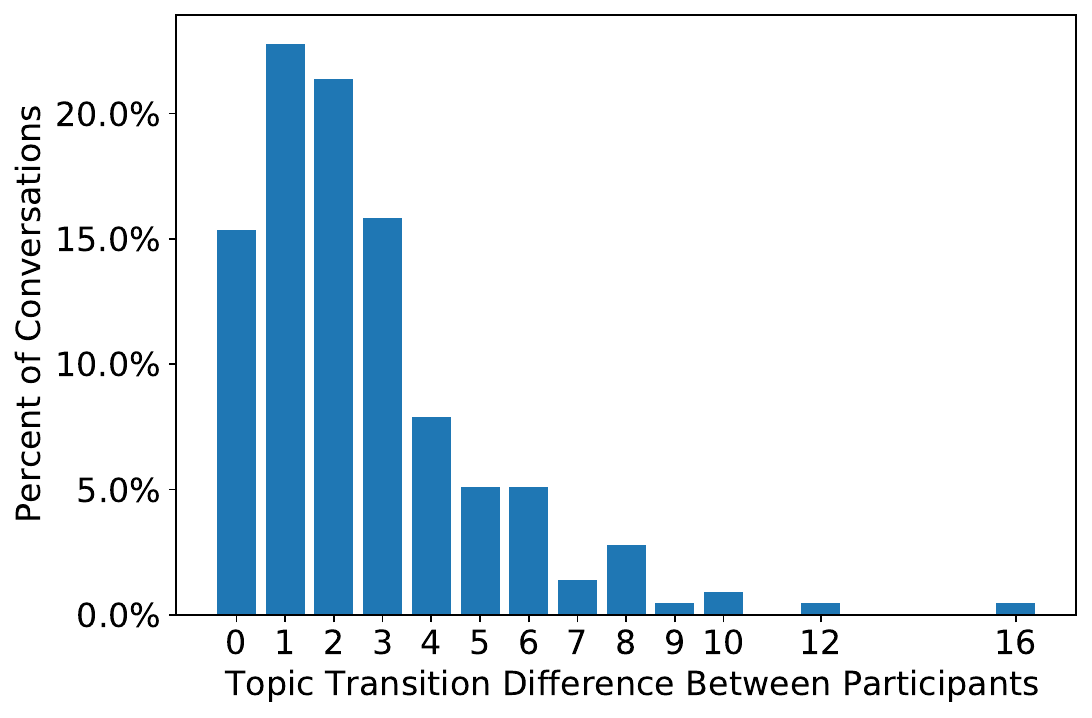}
     \caption{Topic transition difference and percentage of conversations with such difference }
     \end{flushleft}
 \end{figure}
 
\subsection{Time Series Analysis of Topic transition} Next, the study investigate the distribution of utterances per topic as the conversation progresses. Mean and standard deviation of turns/topic is computed for all conversations. It is observed that standard deviation from mean of number of utterances is significant for all topics within a conversation. Hence, we use median to construct a line chart as median is a better measure of central tendency when there are outliers. The correlation between topic time series and number of utterances is observed to be $0.21$ signifying only a weak correlation. Thus, this study did not find any pattern topic transition time series and number of utterances ($(\tilde{\chi}^2 = 11.27, p = 0.98)$).

\begin{figure} [htbp!]
     \begin{flushleft}
     \includegraphics[width=0.49\textwidth]{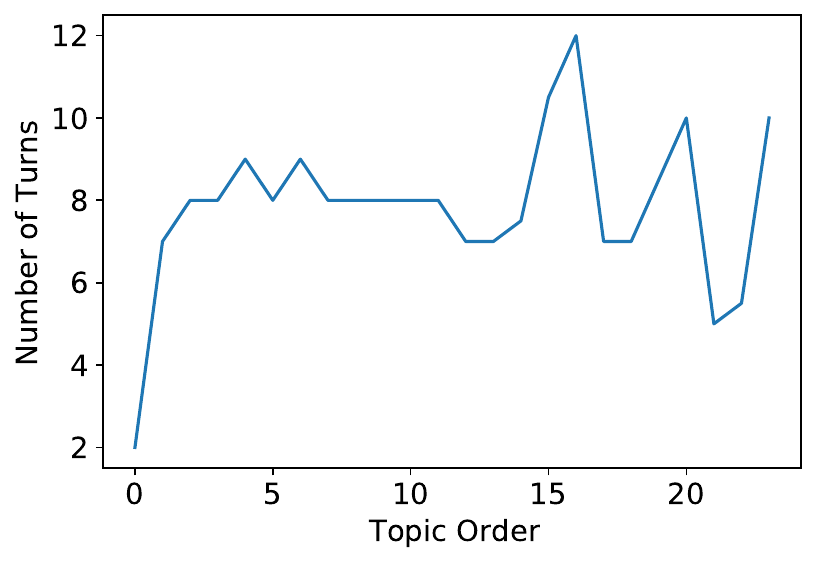}
     \caption{Line plot of turns/topic in a conversation}
     \end{flushleft}
 \end{figure}

\section{Modelling Topic Transition}
In addition to the empirical studies performed, we also modelled topic transition on the manually annotated switchboard corpus, described in section 4.2. Before describing the modelling in detail, we briefly describe the approaches to model topic transition in literature.

Approaches to topical segmentation in dialogue include unsupervised and supervised methods. Unsupervised algorithm work on finding similarity or dissimilarity between segments of text, TextTiling \citep{hearst1997texttiling} is a seminal work in unsupervised topic segmentation. Supervised approaches work with hand-crafted features or deep learning based methods such as used \citep{Arguello_Rose_2006, xing2021improving, konigari2021topic}. 

Following the related research work, we formulate topic transition turn detection as formulated as a binary classification problem. We implement TextTiling \citet{hearst1997texttiling} as a baseline and then proceed to implement classical machine learning as well as deep learning based classification algorithms


Since, dialogue is inherently context based \emph{i.e.} the next utterance is influenced by previous utterances and a topic can span across multiple turns, consecutive utterances are grouped by speaker and termed \emph{turn}. A chunk could consist of multiple utterances where the first utterance might be a participant \textquotesingle s contribution of the current topic, the second utterance could be the introduction of a new topic and a third utterance could be continuation of the newly introduced topic. Turns could also be single utterance where the user did not change topic. 




 TextTiling\citep{hearst1997texttiling} is implemented(Using the code from NLTK \citet{10.5555/1717171}). Turns are formatted as paragraphs separated by two line breaks $(\textbackslash n \textbackslash n)$ as required by TextTiling which works with Lexical Cohesion. The last turn of a paragraph, obtained from Texttiling, is labelled as topic transition turn and all other turns are labelled as topic continuation turns. Additionally, as classic machine learning classifiers, Naive Bayes and LightGBM are implemented.

Finally, utilizing modern deep-learning based classification algorithms, XLNet \citep{yang2019xlnet} is implemented. XLNet is state-of-the-art in text classification tasks \citep{minaee2020deep}. The implementation was based on Hugging Face's Transformers \citep{wolf2019huggingface}. XLNet-base is fine-tuned with $4$ epochs using AdamW (Adam with weight decay) optimizer with Learning Rate of $1e-5$. More than $4$ epochs reduce the train error rate but the difference in valid and train error rate increases. The fine-tuning was done on a single GPU. One epoch took about $28$ minutes to complete.

The performance of algorithms is evaluate against macro averaged precision, recall and f1 score. Precision is a metric indicating how accurately topic transition turn is detected and the values obtained can be seen in Table 2.

\subsection{Results and Error Analysis}

Results in Table 2 show that turns where topic transitions occur can be differentiated from chunks where topics are continued. Evaluation is performed on two test sets. An in-domain $id$ test set which is a subset of annotated switchboard corpus (described in section 4.2) and an out-of-domain($ood$) test set. out-of-domain ($ood$).

It is observe from the study that TextTiling \citep{hearst1997texttiling} is more suitable for expository text since it works with Lexical Cohesion and requires input text to be in paragraphs, which is a property of expository text and not necessarily of a text conversation. Previous studies \citet{konigari2021topic} have also demonstrated that TextTiling \citep{hearst1997texttiling} is more suitable for text with clearly defined topics. In terms of precision, LightGBM performs better than other algorithms for $id$ dataset with a precision of $0.91$. In terms of recall and f1 score, XLNet-base performs better than other algorithms for $id$ test set.

We further evaluate the performance of XLNet (fine-tuned on annotated switchboard corpus) on an out-of-domain ($ood$) test set. The test set selected for $ood$ evaluation is Topical-Chat \citep{Gopalakrishnan2019} as it is also a knowledge-grounded dataset. $10$ conversations from Topical-Chat \citep{Gopalakrishnan2019} are annotated following the annotation guidelines mentioned in section 4.2. Statistic on the $ood$ can be observed in Table 3. Results of evaluation on Topical-chat can be seen in Table 2 with suffix $ood$. It is observed that XLNet-base predicted topic transitions with a precision of $0.78$ on $ood$ test set. TextTiling detects topic transitions with a precision of $0.57$ on $ood$ test set. XLNet performs better overall for $ood$ test set.

\newcolumntype{M}[1]{>{\raggedleft\arraybackslash}m{#1}}
\begin{table}[]
    \begin{centering}
    \begin{tabular}{|m{4.9em}|m{0.6cm}|m{1.32cm}|m{1cm}|m{0.5cm}|} 
    \hline
    \textbf{Model} 
    & \textbf{split}
    & \textbf{Precision} 
    & \textbf{Recall} 
    & \textbf{F1}
    \\ \hline
    Naive Bayes  & $id$ & 0.55 & 0.57 & 0.40\\
    
    LightGBM & $id$ & \textbf{0.91}& 0.50 & 0.46 \\
    \hline
    TextTiling  & $id$ & 0.58 &  0.59 & 0.58\\
    TextTiling  & $ood$ & 0.57 &  0.54 & 0.52\\
    \hline
    XLNet-base & $id$ & 0.68 & \textbf{0.61} & \textbf{0.62}\\
    XLNet-base & $ood$ & \textbf{0.78} & \textbf{0.61} & \textbf{0.59} \\
    \hline
    \end{tabular}
    \caption{Evaluation scores for various algorithms on testset }
    \end{centering}
\end{table}

\newcolumntype{M}[1]{>{\raggedright\arraybackslash}m{#1}}
\begin{table}[htbp!]
\scalebox{1}{
    \footnotesize
    \raggedleft
    \begin{tabular}{
    M{0.7\linewidth} |                
    M{0.1\linewidth}}

    \\
{No. conversations} & 10
 \\ \hline \\ 
{No. total turns} & 216
 \\ \hline \\ 
{Avg. turns per conversation} & 21.6
 \\ \hline \\
{Avg. topics transitions per conversation} & 7.60
 \\ \hline \\
{Avg. turns per topic transitions} & 2.82
 \\
 
\end {tabular}}
\caption{$ood$ testset statistics}
\end{table}

\section{Limitations and Future Work}
Future work will include the application of insights derived from empirical studies to apply them in open-domain dialogue systems such as using the topic transition trained to re-rank responses on topicality. A limitation of this work is the inter-annotator agreement could only be obtained on a small sample of conversation. Future work will include obtaining inter-annotator agreement for all $215$ Switchboard and $10$ Topical-Chat corpora.

\section{Conclusion}

Empirical study on how participants engage in topic transitions in a dyad is presented. It is observed that longer conversations have more topic transitions, topic transition is generally carried out more by one participant and there is no pattern observed in time series of topic transition. Topic transition was also modelled with satisfactory precision and the $ood$ evaluation produced satisfactory results as well.



\bibliography{anthology,custom}

\begin{thebibliography}{34}
\expandafter\ifx\csname natexlab\endcsname\relax\def\natexlab#1{#1}\fi

\bibitem[{Arguello and Rosé(2006)}]{Arguello_Rose_2006}
Jaime Arguello and Carolyn Rosé. 2006.
\newblock Topic-segmentation of dialogue.
\newblock In \emph{Proceedings of the Analyzing Conversations in Text and
  Speech}, page 42–49.

\bibitem[{Bird et~al.(2009)Bird, Klein, and Loper}]{10.5555/1717171}
Steven Bird, Ewan Klein, and Edward Loper. 2009.
\newblock \emph{Natural Language Processing with Python}, 1st edition.
\newblock O’Reilly Media, Inc.

\bibitem[{Bonin et~al.(2012)Bonin, Campbell, and
  Vogel}]{Bonin_Campbell_Vogel_2012}
Francesca Bonin, Nick Campbell, and Carl Vogel. 2012.
\newblock \href {https://doi.org/10.1109/CogInfoCom.2012.6422056} {Laughter and
  topic changes: Temporal distribution and information flow}.
\newblock In \emph{2012 IEEE 3rd International Conference on Cognitive
  Infocommunications (CogInfoCom)}, pages 53--58. IEEE.

\bibitem[{Boufaden et~al.(2001)Boufaden, Lapalme, and
  Bengio}]{Boufaden_Lapalme_Bengio_2001}
Narjès Boufaden, Guy Lapalme, and Yoshua Bengio. 2001.
\newblock Topic segmentation: A first stage to dialog-based information
  extraction.
\newblock In \emph{In Natural Language Processing Pacific Rim Symposium,
  NLPRS’01}. Citeseer.

\bibitem[{Brown and Yule(1983)}]{Brown_Yule_1983}
Gillian Brown and George Yule. 1983.
\newblock \emph{Discourse Analysis}.
\newblock Cambridge University Press.
\newblock Google-Books-ID: ZUnEAgAAQBAJ.

\bibitem[{Bublitz(1988)}]{Bublitz_1988}
Wolfram Bublitz. 1988.
\newblock \emph{Supportive Fellow-speakers and Cooperative Conversations:
  Discourse Topics and Topical Actions, Participant Roles and ``Recipient
  Action" in a Particular Type of Everyday Conversation}.
\newblock John Benjamins Publishing.
\newblock Google-Books-ID: d85TIjf7odQC.

\bibitem[{Calhoun et~al.(2010)Calhoun, Carletta, Brenier, Mayo, Jurafsky,
  Steedman, and
  Beaver}]{Calhoun_Carletta_Brenier_Mayo_Jurafsky_Steedman_Beaver_2010}
Sasha Calhoun, Jean Carletta, Jason~M. Brenier, Neil Mayo, Dan Jurafsky, Mark
  Steedman, and David Beaver. 2010.
\newblock The nxt-format switchboard corpus: a rich resource for investigating
  the syntax, semantics, pragmatics and prosody of dialogue.
\newblock \emph{Language resources and evaluation}, 44(4):387--419.

\bibitem[{Crow(1983)}]{Crow_1983}
B~Crow. 1983.
\newblock \emph{Topic shifts in couples' conversations.} SAGE Publications,
  Inc.

\bibitem[{Galley et~al.(2003)Galley, McKeown, Fosler-Lussier, and
  Jing}]{Galley_McKeown_Fosler_Lussier_Jing_2003}
Michel Galley, Kathleen McKeown, Eric Fosler-Lussier, and Hongyan Jing. 2003.
\newblock \href {https://doi.org/10.3115/1075096.1075167} {Discourse
  segmentation of multi-party conversation}.
\newblock In \emph{Proceedings of the 41st Annual Meeting on Association for
  Computational Linguistics - Volume 1}, ACL ’03, page 562–569. Association
  for Computational Linguistics.
\newblock Event-place: Sapporo, Japan.

\bibitem[{Garcia and Joanette(1997)}]{Garcia_Joanette_1997}
Linda~J. Garcia and Yves Joanette. 1997.
\newblock \href {https://doi.org/10.1006/brln.1997.1871} {Analysis of
  conversational topic shifts: A multiple case study}.
\newblock \emph{Brain and Language}, 58(1):92--114.

\bibitem[{Gardner(1984)}]{Gardner_1984}
Roderick Gardner. 1984.
\newblock \href {https://doi.org/10.1017/S0261444800010545} {Discourse
  analysis: implications for language teaching, with particular reference to
  casual conversation}.
\newblock \emph{Language Teaching}, 17(2):102--117.

\bibitem[{Godfrey and Holliman(1993)}]{Godfrey_Holliman_1993}
John~J. Godfrey and Edward Holliman. 1993.
\newblock \href {https://catalog.ldc.upenn.edu/LDC97S62} {Switchboard-1 release
  2 - {LDC}97s62 - linguistic data consortium}.

\bibitem[{Gopalakrishnan et~al.(2019)Gopalakrishnan, Hedayatnia, Chen,
  Gottardi, Kwatra, Venkatesh, Gabriel, and Hakkani-Tür}]{Gopalakrishnan2019}
Karthik Gopalakrishnan, Behnam Hedayatnia, Qinlang Chen, Anna Gottardi, Sanjeev
  Kwatra, Anu Venkatesh, Raefer Gabriel, and Dilek Hakkani-Tür. 2019.
\newblock \href {https://doi.org/10.21437/Interspeech.2019-3079}
  {{Topical-Chat: Towards Knowledge-Grounded Open-Domain Conversations}}.
\newblock In \emph{Proc. Interspeech 2019}, pages 1891--1895.

\bibitem[{Hearst(1997)}]{hearst1997texttiling}
Marti~A Hearst. 1997.
\newblock Texttiling: Segmenting text into multi-paragraph subtopic passages.
\newblock \emph{Computational linguistics}, 23(1):33--64.

\bibitem[{Hsueh et~al.(2006)Hsueh, Moore, and Renals}]{Hsueh_Moore_Renals_2006}
Pei-Yun Hsueh, Johanna~D. Moore, and Steve Renals. 2006.
\newblock \href {https://www.aclweb.org/anthology/E06-1035} {Automatic
  segmentation of multiparty dialogue}.
\newblock In \emph{11th Conference of the European Chapter of the Association
  for Computational Linguistics}. Association for Computational Linguistics.

\bibitem[{Joty et~al.(2011)Joty, Carenini, Murray, and Ng}]{joty2011supervised}
Shafiq Joty, Giuseppe Carenini, Gabriel Murray, and Raymond~T Ng. 2011.
\newblock Supervised topic segmentation of email conversations.
\newblock In \emph{Fifth International AAAI Conference on Weblogs and Social
  Media}.

\bibitem[{Konigari et~al.(2021)Konigari, Ramola, Alluri, and
  Shrivastava}]{konigari2021topic}
Rachna Konigari, Saurabh Ramola, Vijay~Vardhan Alluri, and Manish Shrivastava.
  2021.
\newblock Topic shift detection for mixed initiative response.
\newblock In \emph{Proceedings of the 22nd Annual Meeting of the Special
  Interest Group on Discourse and Dialogue}, pages 161--166.

\bibitem[{Lambrecht(1996)}]{Lambrecht_1996}
Knud Lambrecht. 1996.
\newblock \emph{Information Structure and Sentence Form: Topic, Focus, and the
  Mental Representations of Discourse Referents}.
\newblock Cambridge University Press.
\newblock Google-Books-ID: bsXLCgAAQBAJ.

\bibitem[{Mann et~al.(1977)Mann, Carlisle, Moore, and
  Levin}]{Mann_Carlisle_Moore_Levin_1977}
William~C. Mann, James~H. Carlisle, James~A. Moore, and James~A. Levin. 1977.
\newblock \href {https://eric.ed.gov/?id=ED136779} {\emph{An Assessment of
  Reliability of Dialogue-Annotation Instructions}}.
\newblock ISI/RR-77-54.

\bibitem[{Maynard(2009)}]{Maynard_2009}
Douglas~W. Maynard. 2009.
\newblock \href {https://doi.org/10.1515/semi.1980.30.3-4.263} {Placement of
  topic changes in conversation}.
\newblock \emph{Semiotica}, 30(3-4):263--290.

\bibitem[{Minaee et~al.(2020)Minaee, Kalchbrenner, Cambria, Nikzad, Chenaghlu,
  and Gao}]{minaee2020deep}
Shervin Minaee, Nal Kalchbrenner, Erik Cambria, Narjes Nikzad, Meysam
  Chenaghlu, and Jianfeng Gao. 2020.
\newblock Deep learning based text classification: A comprehensive review.
\newblock \emph{arXiv preprint arXiv:2004.03705}.

\bibitem[{Passonneau and Litman(1997)}]{Passonneau_Litman_1997}
Rebecca~J. Passonneau and Diane~J. Litman. 1997.
\newblock \href {http://dl.acm.org/citation.cfm?id=972684.972689} {Discourse
  segmentation by human and automated means}.
\newblock \emph{Comput. Linguist.}, 23(1):103--139.

\bibitem[{Planalp and Tracy(1980)}]{Planalp_Tracy_1980}
Sally Planalp and Karen Tracy. 1980.
\newblock \href {https://doi.org/10.1080/23808985.1980.11923805} {Not to change
  the topic but...: A cognitive approach to the management of conversation}.
\newblock \emph{Annals of the International Communication Association},
  4(1):237--258.

\bibitem[{Purver et~al.(2011)Purver, Tur, and
  De~Mori}]{Purver_Tur_De_Mori_2011}
Matthew Purver, Gokhan Tur, and Rento De~Mori. 2011.
\newblock \href
  {https://www.wiley.com/en-ie/Spoken+Language+Understanding:+Systems+for+Extracting+Semantic+Information+from+Speech-p-9780470688243}
  {\emph{Topic segmentation}}, page 291–317. John Wiley `|\&' Sons.

\bibitem[{Reynar(1994)}]{reynar1994automatic}
Jeffrey~C Reynar. 1994.
\newblock An automatic method of finding topic boundaries.
\newblock \emph{arXiv preprint cmp-lg/9406017}.

\bibitem[{Ries(2001)}]{Ries_2001}
Klaus Ries. 2001.
\newblock \href {https://doi.org/10.1007/3-540-45637-6_5} {Segmenting
  conversations by topic, initiative, and style}.
\newblock In \emph{Information Retrieval Techniques for Speech Applications},
  Lecture Notes in Computer Science, pages 51--66. Springer, Berlin,
  Heidelberg.

\bibitem[{Riou(2015)}]{riou2015methodology}
Marine Riou. 2015.
\newblock A methodology for the identification of topic transitions in
  interaction.
\newblock \emph{Discours. Revue de linguistique, psycholinguistique et
  informatique. A journal of linguistics, psycholinguistics and computational
  linguistics}, (16).

\bibitem[{Sapru and Bourlard(2014)}]{Sapru_Bourlard_2014}
Ashtosh Sapru and Hervé Bourlard. 2014.
\newblock \href
  {http://193.6.4.39/~czap/letoltes/IS14/IS2014/PDF/AUTHOR/IS141199.PDF}
  {Detecting speaker roles and topic changes in multiparty conversations using
  latent topic models.}
\newblock In \emph{INTERSPEECH}, page 2882–2886.

\bibitem[{Sevegnani et~al.(2021)Sevegnani, Howcroft, Konstas, and
  Rieser}]{sevegnani2021otters}
Karin Sevegnani, David~M Howcroft, Ioannis Konstas, and Verena Rieser. 2021.
\newblock Otters: One-turn topic transitions for open-domain dialogue.
\newblock \emph{arXiv preprint arXiv:2105.13710}.

\bibitem[{Van~Dijk(1977)}]{Van_Dijk_1977}
Teun~A. Van~Dijk. 1977.
\newblock \href
  {http://discourses.org/OldArticles/Sentence%20topic%20and%20discourse%20topic.pdf}
  {Sentence topic and discourse topic}.
\newblock \emph{Papers in slavic philology}, 1:49--61.

\bibitem[{Wolf et~al.(2019)Wolf, Debut, Sanh, Chaumond, Delangue, Moi, Cistac,
  Rault, Louf, Funtowicz et~al.}]{wolf2019huggingface}
Thomas Wolf, Lysandre Debut, Victor Sanh, Julien Chaumond, Clement Delangue,
  Anthony Moi, Pierric Cistac, Tim Rault, R{\'e}mi Louf, Morgan Funtowicz,
  et~al. 2019.
\newblock Huggingface's transformers: State-of-the-art natural language
  processing.
\newblock \emph{ArXiv}, pages arXiv--1910.

\bibitem[{Xing and Carenini(2021)}]{xing2021improving}
Linzi Xing and Giuseppe Carenini. 2021.
\newblock Improving unsupervised dialogue topic segmentation with
  utterance-pair coherence scoring.
\newblock \emph{arXiv preprint arXiv:2106.06719}.

\bibitem[{Yang et~al.(2019)Yang, Dai, Yang, Carbonell, Salakhutdinov, and
  Le}]{yang2019xlnet}
Zhilin Yang, Zihang Dai, Yiming Yang, Jaime Carbonell, Ruslan Salakhutdinov,
  and Quoc~V. Le. 2019.
\newblock \href {http://arxiv.org/abs/1906.08237} {Xlnet: Generalized
  autoregressive pretraining for language understanding}.

\bibitem[{Zhang and Zhou(2019)}]{zhang2019topic}
Leilan Zhang and Qiang Zhou. 2019.
\newblock Topic segmentation for dialogue stream.
\newblock In \emph{2019 Asia-Pacific Signal and Information Processing
  Association Annual Summit and Conference (APSIPA ASC)}, pages 1036--1043.
  IEEE.

\end{thebibliography}
\bibliographystyle{acl_natbib}

\clearpage

\appendix


\section{Utterance Count Per Topic}

In addition to plotting median utterances per topic, we also plot mean, minimum and maximum number of utterances as topic order progresses.

\begin{figure} [htbp!]
     \begin{flushleft}
     \includegraphics[width=0.49\textwidth]{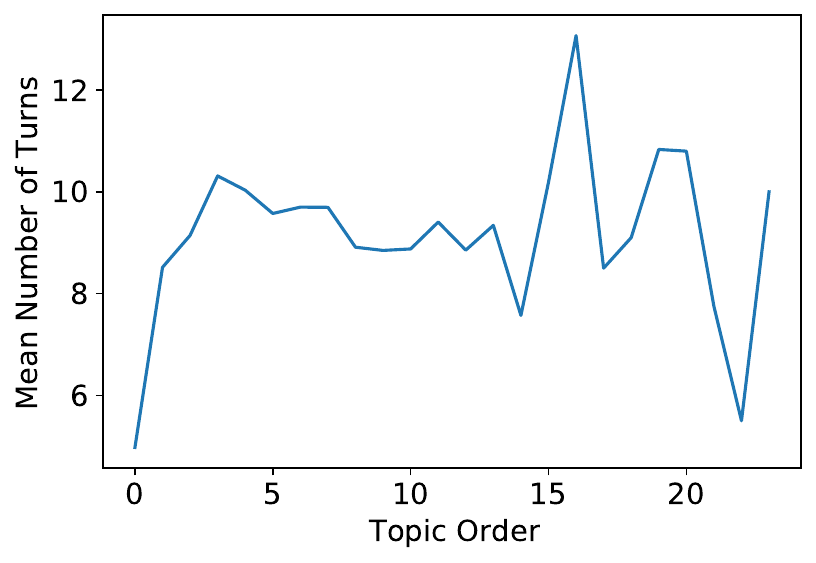}
     \caption{Line plot of mean utterances per topic}
     \end{flushleft}
 \end{figure}
 
 \begin{figure} [htbp!]
     \begin{flushleft}
     \includegraphics[width=0.49\textwidth]{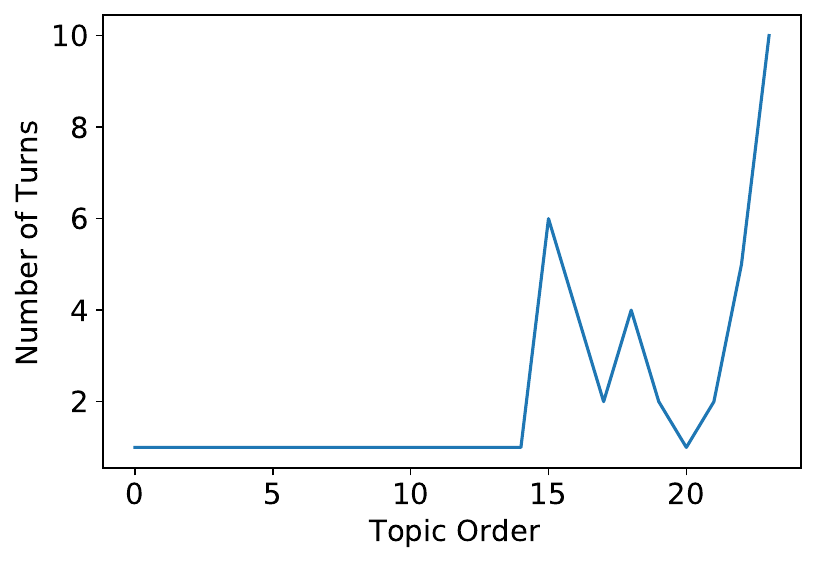}
     \caption{Line plot of minimum utterances per topic}
     \end{flushleft}
 \end{figure}

 \begin{figure} [htbp!]
     \begin{centering}
     \includegraphics[width=0.49\textwidth]{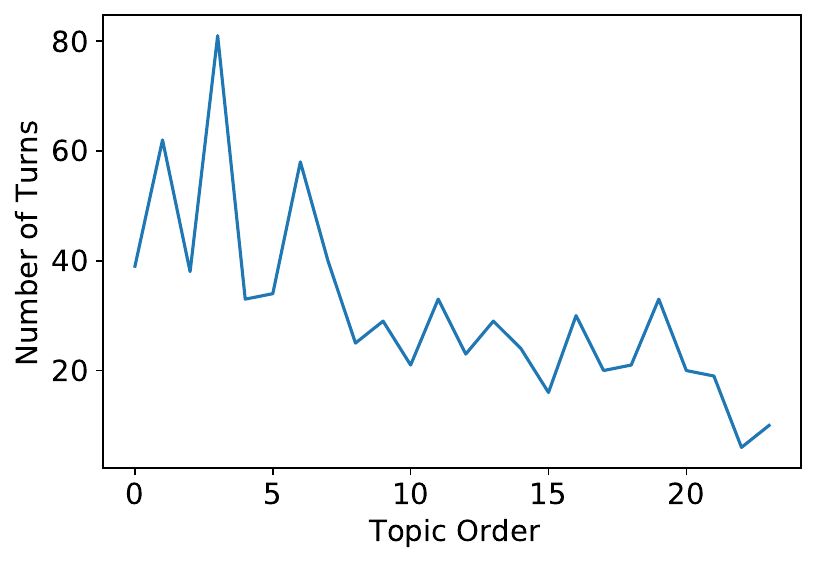}
     \caption{Line plot of maximum utterances per topic}
     \end{centering}
 \end{figure}

\section{t-SNE Visualizations}
To empirically understand the separation of topic transition turns and topic continuation turns, we visualize the two classes using a t-SNE plot.

 \begin{figure} [htbp!]
     \begin{centering}
     \fbox{\includegraphics[width=0.60\textwidth]{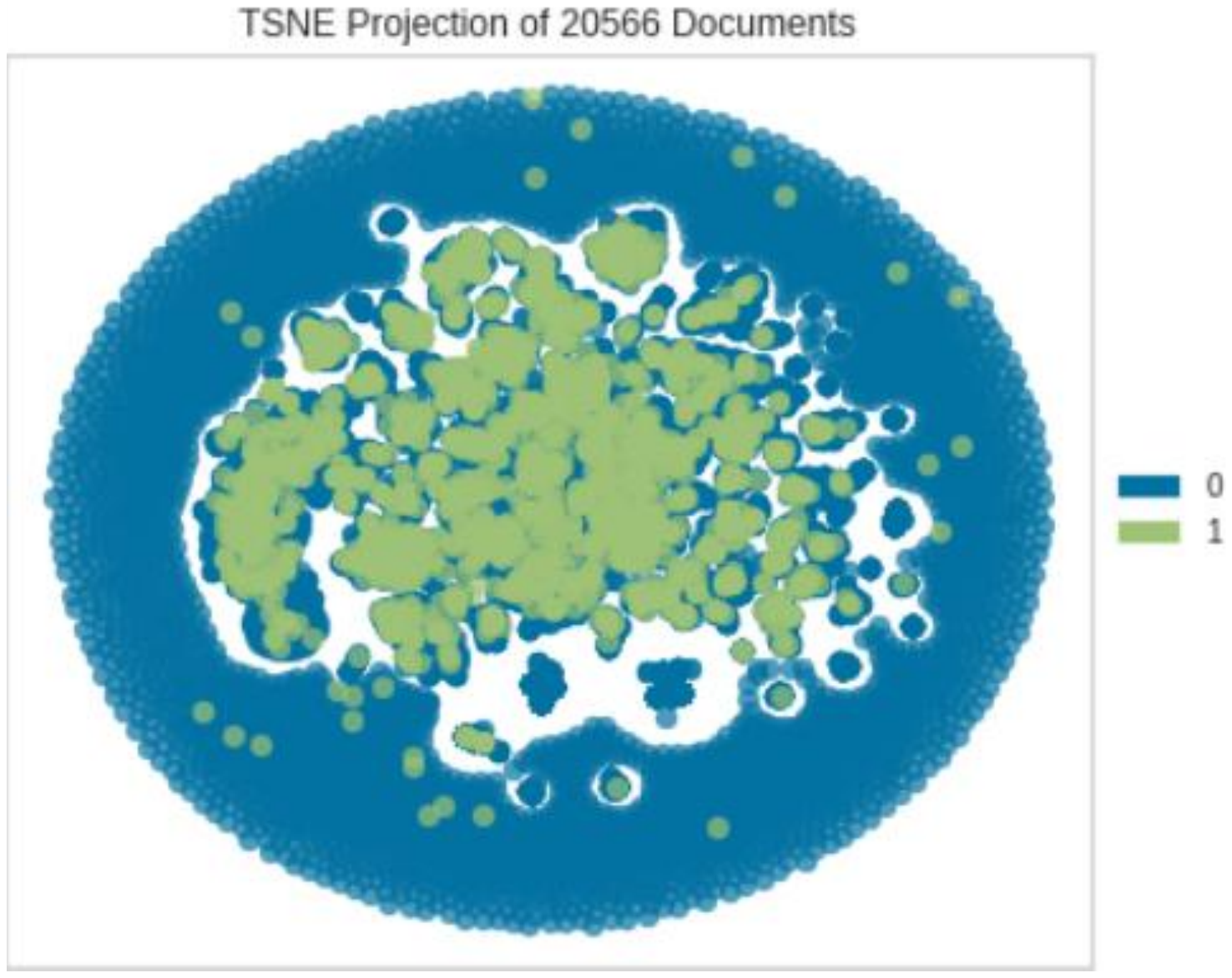}}
     \caption{t-SNE visualization of utterances}
     \end{centering}
 \end{figure}

\clearpage

\end{document}